\pdfoutput=1

\documentclass[11pt]{article}

\usepackage[final]{coling}

\usepackage{times}
\usepackage{latexsym}

\usepackage[T1]{fontenc}

\usepackage[utf8]{inputenc}

\usepackage{microtype}

\usepackage{inconsolata}

\usepackage{graphicx}
\usepackage{algorithm}
\usepackage{algorithmic}
\usepackage{tabularx}
\usepackage{amsmath}
\usepackage{graphicx}  
\usepackage{subcaption} 
\usepackage{amsfonts}
\usepackage{csquotes} 
\usepackage{hyperref}
\usepackage{authblk}

\usepackage{threeparttable}  
\usepackage{booktabs} 
\usepackage{multirow} 

\usepackage{booktabs}
\usepackage{adjustbox}

\title{Unveiling Uncertainty: A Deep Dive into Calibration and Performance of Multimodal Large Language Models
\thanks{The first two authors contributed equally to this work. Corresponding author: Wenbo Hu (\href{mailto:wenbohu@hfut.edu.cn}{wenbohu@hfut.edu.cn.})}
}

\author[1,2]{Zijun Chen}
\author[1]{Wenbo Hu}
\author[3]{Guande He}
\author[4]{Zhijie Deng}
\author[2]{Zheng Zhang}
\author[1]{Richang Hong}

\affil[1]{Hefei University of Technology, Hefei, China}
\affil[2]{Data Space Research Institute, Hefei, China}
\affil[3]{UT Austin, Austin, USA}
\affil[4]{Shanghai Jiao Tong University, Shanghai, China}

\begin{document}
\maketitle
\begin{abstract}
Multimodal large language models (MLLMs) combine visual and textual data for tasks such as image captioning and visual question answering. Proper uncertainty calibration is crucial, yet challenging, for reliable use in areas like healthcare and autonomous driving.
This paper investigates representative MLLMs, focusing on their calibration across various scenarios, including before and after visual fine-tuning, as well as before and after multimodal training of the base LLMs. We observed miscalibration in their performance, and at the same time, no significant differences in calibration across these scenarios. We also highlight how uncertainty differs between text and images and how their integration affects overall uncertainty.
To better understand MLLMs' miscalibration and their ability to self-assess uncertainty, we construct the IDK (I don't know) dataset, which is key to evaluating how they handle unknowns. Our findings reveal that MLLMs tend to give answers rather than admit uncertainty, but this self-assessment improves with proper prompt adjustments. Finally, to calibrate MLLMs and enhance model reliability, we propose techniques such as temperature scaling and iterative prompt optimization. Our results provide insights into improving MLLMs for effective and responsible deployment in multimodal applications. Code and IDK dataset: \href{https://github.com/hfutml/Calibration-MLLM}{https://github.com/hfutml/Calibration-MLLM}.
\end{abstract}

\section{INTRODUCTION}
Multimodal large language models (MLLMs) represent a significant advancement in artificial intelligence by merging the capabilities of processing both visual and textual inputs~\cite{khan2021exploiting,lei2021understanding,radford2021learning,641130e378d68457a4a2986f,liu2024valor}. These models, trained on large datasets containing paired images and texts, excel in tasks such as image captioning, visual question answering, and cross-modal retrieval by correlating visual elements with corresponding textual descriptions.

However, like LLMs, MLLMs also suffer from issues such as hallucinations and unreliability \cite{bai2024hallucination}. To identify and address these problems, quantifying and then calibrating the uncertainty of these models is an important approach. This process aligns the model's uncertainty with its actual predictive accuracy, similar to how humans can roughly evaluate their confidence in a particular matter. Accurate calibration is essential for MLLMs in safety-critical fields like autonomous driving \cite{yang2023real}, medical care \cite{buddenkotte2023calibrating}, drug discovery~\cite{li2024conformalized} and weather prediction \cite{price2024probabilistic} to avoid overconfidence that can compromise reliability. Overconfidence in model predictions can lead to false assurances, increasing the risk of catastrophic errors, such as misdiagnoses or inaccurate weather forecasts, ultimately endangering lives and safety.

Recent studies have raised concerns about model fine-tuning and alignment of large language models inducing overconfidence~\cite{kadavath2022language,tian2023just}, leading to suboptimal calibration~\cite{641130e378d68457a4a2986f,he2023investigating,649e52bfd68f896efae47fa7}. These works have shown notable success in identifying and addressing overconfidence in unimodal models using calibration techniques, thereby improving the reliability of model predictions in single-modality tasks. Despite these advancements, the effect of miscalibration on MLLMs, remains underexplored. Proper calibration of MLLMs unlocks their full potential, enabling reliable use across various applications~\cite{murphy1977reliability}, especially in more rich image-text settings. Compared with unimodal models, MLLM needs multimodal information input, making the analysis of model confidence more complex and challenging. Whether there is a difference in the confidence for different modal information and how the model confidence changes during the integration of different modal information are all worthy of exploration.
 
Our research investigates the uncertainty calibration of state-of-the-art MLLMs, such as LLaVA~\cite{643e0ad60746dc40e341a515} and Qwen-VL~\cite{6578f5ee939a5f4082a262c4}. We first observed the calibration differences in several models across various settings, specifically focusing on before and after fine-tuning, and performance of linguistic tasks (compared to their corresponding base LLM). While the calibration differences of this two scenarios, were not substantial, we found that, overall, MLLMs consistently exhibited miscalibration. We then delved deeper into uncertainty quantification for MLLMs, exploring the differences in uncertainty exhibited by MLLMs when processing information from images versus text, and examining how the continuous integration of these two modalities affects uncertainty. Following this, we constructed the IDK dataset to further investigate the overconfidence issue in MLLMs and assess whether it can be mitigated with simple prompts. Finally, we introduced several calibration techniques to achieve more accurate calibration of MLLMs.

The contribution of this paper can be summarized as follows:
\begin{itemize}
    \item \textbf{Calibration stability, yet persistent miscalibration:} We show that MLLMs maintain relatively consistent calibration before and after fine-tuning, alleviating concerns of degraded calibration. Our findings also suggest that MLLMs have minimal impact on the calibration of linguistic tasks after training from the base LLM, enabling visual data integration without significantly affecting original linguistic capabilities. However, the overall calibration of MLLMs is still suboptimal.
    \item \textbf{Differentiated uncertainty integration:} We observed that compared to images, MLLMs have lower uncertainty in the information of text, and the information of the two modalities can be integrated together to reduce the uncertainty of the model.
    \item \textbf{IDK Dataset and OOD assessment:} We constructed the IDK dataset by having the model repeatedly answer multiple times and created the OOD (out of distribution) dataset using recent news and GPT-3.5. MLLMs often answer questions even when unsure, but prompt-based encouragement can mitigate this, as seen in OOD data.
    \item \textbf{Advanced calibration techniques:} We propose and validate advanced calibration strategies, including temperature scaling and iterative prompt optimization, which significantly improve the reliability and effectiveness of MLLMs in diverse applications.
\end{itemize}

\section{PRELIMINARIES}
In this section, we briefly introduce the basics and the training process of MLLMs, followed by an explanation of how to quantify uncertainty under the multiple-choice setting.

\subsection{Multimodal Large Language Models}
With the rise of LLMs, many works have begun to focus on MLLMs. Most MLLMs are retrained by adding an image encoding part to LLMs~\cite{643e0ad60746dc40e341a515}. This gives LMMs the ability to process vision. \citet{643e0ad60746dc40e341a515} proposed a new training method for multimodal models: training through instruction-following data.  \citet{sun2023aligning} argued that MLLMs are constructed across multiple modalities, and the misalignment between the two modalities may lead to ``hallucination''. Previous work has explored the issue of calibration in LMs. \citet{kadavath2022language} demonstrated that advanced large-scale pretrained language models exhibit good calibration, while aligned language models (using fine-tuning or reinforcement learning to align human usage habits) are poorly calibrated due to being overly confident in logit-based multiple-choice questions. In the field of uncertainty, some researchers have also made significant contributions. They have decomposed the uncertainty of models into finer components. For instance, \citet{kendall2017uncertainties} and \citet{malinin2018predictive} decomposed uncertainty into model uncertainty and data uncertainty, while \citet{he2023investigating} proposed the existence of format uncertainty and answer uncertainty in the logit-based multiple-choice setting. Their work provides new insights and approaches for more rational and accurate assessment of model calibration.

\subsection{The Training Process of MLLMs}

In this section, we will use LLaVA as an example to illustrate its training process.

\textbf{Stage 1: Pre-training for Feature Alignment.}
During training, LLaVA maintain the visual encoder and LLM weights in a frozen state, focusing on maximizing the likelihood with only trainable parameters represented by the projection matrix. This approach allows for aligning the image features with the pre-trained LLM word embedding. Essentially, this stage can be viewed as training a compatible visual tokenizer specifically designed for the frozen LLM.

\textbf{Stage 2: Fine-tuning End-to-End.} Always keep the visual encoder weights frozen, and continue to update both the pre-trained weights of the projection layer and LLM in LLaVA. To enhance LLaVA's ability to follow instructions more effectively, the authors fine-tuned the model using language-image instruction-following dataset.

\subsection{Uncertainty Quantification and Calibration of MLLMs}
Uncertainty quantification and calibration in machine learning models are crucial for evaluating the alignment between a model's confidence in its predictions and the actual accuracy of those predictions~\cite{guo2017calibration,kendall2017uncertainties,cui2020calibrated,liu2023deep}. Misalignment, where a model is overly confident in incorrect predictions or uncertain about accurate ones, undermines its reliability and trustworthiness, especially in real-world applications that depend on the model's certainty for decision-making. Recent studies have seen the development and use of various effective uncertainty quantification methods, such as logit-based likelihood \cite{he2023investigating}, semantic entropy \cite{kuhn2023semantic}, and self-expression \cite{xiong2023can}. These approaches offer valuable insights for assessing uncertainty in both LLMs and MLLMs.

To assess how well a model’s confidence or uncertainty aligns with its accuracy, Expected Calibration Error (ECE) serves as a key metric. The goal of ECE is to quantify the difference between a model's predicted confidence and its actual performance. This is done by dividing predictions into bins, calculating the average accuracy and confidence for each bin, and summing the weighted differences across all bins:

\begin{equation}
\text{ECE} = \sum_{m=1}^{M} \frac{N_m}{N} \left|\text{acc}(B_m) - \text{conf}(B_m)\right|,
\end{equation}
where \(M\) is the number of bins, \(N\) is the total number of samples, \(N_m\) is the number of samples in the bin \(B_m\), and \(\text{acc}(B_m)\) and \(\text{conf}(B_m)\) represent the actual accuracy and predicted confidence in the bin \(B_m\). A smaller ECE indicates better model calibration. We use 10 equal-sized bins for ECE calculation.

We also use two ECE variants:
\begin{itemize}
    \item \textbf{Maximum Calibration Error (MCE):} the maximum difference between accuracy and confidence across all bins. This provides a measure of the worst-case scenario that model calibration errors may reach.
    \[
    \text{MCE} = \max_{m=1}^{M} |\text{acc}(B_m) - \text{conf}(B_m)|
    \]
    
    \item \textbf{Normalized Expected Calibration Error (ENCE):} normalizes the error by dividing by the predicted confidence of each bin, giving more balanced calibration across varying confidence levels.
    \[
    \text{ENCE} = \sum_{m=1}^{M} \frac{N_m}{N} \frac{|\text{acc}(B_m) - \text{conf}(B_m)|}{\text{conf}(B_m)}
    \]
\end{itemize}

\section{ANALYSIS OF CALIBRATION ACROSS DIFFERENT SCENARIOS}

In this section, we adopt logits-based likelihood as the uncertainty quantification method and primarily examine whether there are significant differences in the calibration of MLLMs at different scenarios. Additionally, we aim to observe if any miscalibration occurs.

\subsection{Calibration Differences Between Before and After Fine-tuning MLLMs}
In LLMs, some studies have shown that LLMs have good calibration during the Pre-Trained phase, but their calibration decreases after fine-tuning \cite{he2023investigating}. To demonstrate how the fine-tuning process in MLLMs might influence calibration performance, we use some VQA (Visual Question Answering) datasets to test calibration of them at different stages. 

\textbf{Experiment Setup: } \\
\textbf{Model: }The pre-trained MLLMs we selected include the LLaVA-v1.5-7B~\cite{643e0ad60746dc40e341a515} and LLaVA-v1.5-13B trained after Stage 1 (without fine-tuning), and Qwen-VL. The corresponding aligned MLLMs for these models are the LLaVA trained after Stage 2 and Qwen-VL-Chat~\cite{6578f5ee939a5f4082a262c4} \\
\textbf{Dataset: }We selected same VQA task datasets covering a wide range, including MMBench \cite{MMBench}, SEED-Bench \cite{64c88ca43fda6d7f062687f8}, BilbaoQA2~\cite{BilbaoQA22023}, RealworldQA~\cite{realworldqa2024}, MathVers~\cite{zhang2024mathverse}, Creature (biological habits generated by GPT-4), and ScreenShot (manually annotated screenshots from films). These datasets involve evaluations of various abilities such as perception and reasoning. We randomly sample some subsets from these datasets for testing. 

\textbf{Stability of Calibration After Fine-tuning: }
As can be seen in Table~\ref{tab:mmbench}, we list the accuracy, confidence and ECE (MCE and ENCE) of the MLLMs before and after fine-tuning, namely the stage 1 and stage 2 (The remaining data results are shown in the Appendix A.) We observed that after fine-tuning, the accuracy and the average confidence of the model improved on various datasets, this is similar to LLMs~\cite{kadavath2022language,he2023investigating}. However, the changes in ECE (MCE and ENCE) were inconsistent. In some datasets, calibration decreased after fine-tuning, while in others, it improved, with the latter being more common. This suggests that calibration changes in MLLMs may not be directly tied to fine-tuning.

\begin{table}[h!]
\centering
\scriptsize 
\begin{tabular}{lcccccc}
\toprule
Model & Acc & Conf & ECE & MCE & ENCE \\
\midrule
LLaVA-7B-Stage1 & 0.542 & 0.448 & 0.118 & 0.34 & 0.234 \\

LLaVA-7B-Stage2 & 0.741 & 0.757 & \textbf{0.075} & \textbf{0.246} & \textbf{0.11} \\
\midrule
LLaVA-13B-Stage1 & 0.636 & 0.423 & 0.235 & \textbf{0.383} & 0.306 \\
LLaVA-13B-Stage2 & 0.741 & 0.836 & \textbf{0.101} & 0.387 & \textbf{0.181} \\
\midrule
Qwen-VL & 0.742 & 0.604 & 0.137 & 0.28 & 0.196 \\
Qwen-VL-Chat & 0.762 & 0.755 & \textbf{0.079} & \textbf{0.193} & \textbf{0.151} \\
\bottomrule
\end{tabular}
\caption{ECE (MCE and ENCE), accuracy, and confidence of several MLLMs tested on MMBench}
\label{tab:mmbench}
\end{table}

\subsection{Calibration Changes in MLLMs for Linguistic Tasks}

Many MLLMs like LLaVA, update the base LLM weights during training. Ideally, these models should retain their original linguistic task calibration after gaining the ability to process images. In this section, we compare several model pairs on linguistic tasks to assess calibration changes.

\textbf{Experiment Setup: } \\
\textbf{Model: }On the model side, we selected several pairs of models for comparison, namely Qwen-VL and Qwen-7B, LLaVA-v1.5-7B and Vicuna-v1.5-7B, LLaVA-v1.5-13B and Vicuna-v1.5-13B, LLaVA-llama-2-13b-chat-lightning-preview \cite{643e0ad60746dc40e341a515} and LLaMA-2-13B-Chat \cite{touvron2023llama} \\
\textbf{Dataset: }The linguistic task datasets we selected include: ARC \cite{allenai:arc}, CommonsenseQA \cite{talmor-etal-2019-commonsenseqa}, MMLU \cite{hendryckstest2021}, OpenBookQA \cite{OpenBookQA2018}, and RACE \cite{lai-etal-2017-race}. We randomly sample some subsets from these datasets for testing. All these datasets are configured in a multiple-choice format.

\begin{table}[h!]
\centering
\scriptsize 
\begin{tabular}{lcccccc}
\toprule
Model & Acc & Conf & ECE & MCE & ENCE \\
\midrule
Vicuna-7B & 0.375 & 0.589 & \textbf{0.213} & \textbf{0.349} & 0.377 \\
LLaVA-7B & 0.421 & 0.640 & 0.224 & 0.705 & \textbf{0.355} \\
\midrule
Vicuna-13B & 0.414 & 0.667 & \textbf{0.253} & \textbf{0.455} & \textbf{0.404} \\
LLaVA-13B & 0.431 & 0.739 & 0.308 & 0.490 & 0.443 \\
\midrule
LLaMA2-13B-Chat & 0.400 & 0.667 & 0.267 & 0.571 & 0.435 \\
LLaVA-LLaMA2 & 0.407 & 0.636 & \textbf{0.229} & \textbf{0.349} & \textbf{0.399} \\
\midrule
Qwen-7B & 0.424 & 0.550 & 0.134 & 0.701 & 0.251 \\
Qwen-VL-Chat & 0.500 & 0.599 & \textbf{0.099} & \textbf{0.224} & \textbf{0.199} \\
\bottomrule
\end{tabular}
\caption{ECE (MCE and ENCE), accuracy, and confidence of several models tested on MMLU, the LLaVA models used here are fully trained models after Stage 2}
\label{tab:mmlu}
\end{table}

\textbf{Minimal Impact of Linguistic Tasks: }
 We report the accuracy and ECE (MCE and ENCE) of the MLLMs in terms of the linguistic QA tasks in Table~\ref{tab:mmlu}. Remaining results are shown in the Appendix B. The results are compared between the uni-modal LLM and multimodal models. It can be seen that, for these datasets, calibration in some cases slightly increases, while in others it slightly decreases. This indicates that the calibration of MLLMs for linguistic tasks did not significantly deteriorate with the update of model parameter weights. 

\textbf{Analysis of Two Experiments: }
Although we did not observe significant calibration differences in MLLMs across different settings in the previous two experiments, we still identified instances of miscalibration in many tasks. For example, in Table~\ref{tab:mmlu}, LLaVA-13B has an accuracy of 0.431, but its confidence reaches 0.739, indicating clear overconfidence. This is evident in several datasets where the model's accuracy and confidence are not aligned, and the ECE (MCE and ENCE) values are often relatively high.

\section{UNCERTAINTY ANALYSIS AND MULTIMODAL INTEGRATION}
In the context of MLLMs, which incorporates the visual domain, the sources of uncertainty become even more diverse, encompassing elements from both textual and visual modalities. 
This line of inquiry not only advances our comprehension of MLLMs' operational dynamics but also highlights the complexity of managing uncertainty in a multimodal setting.

\begin{figure}
    \centering
    \includegraphics[width=1\linewidth]{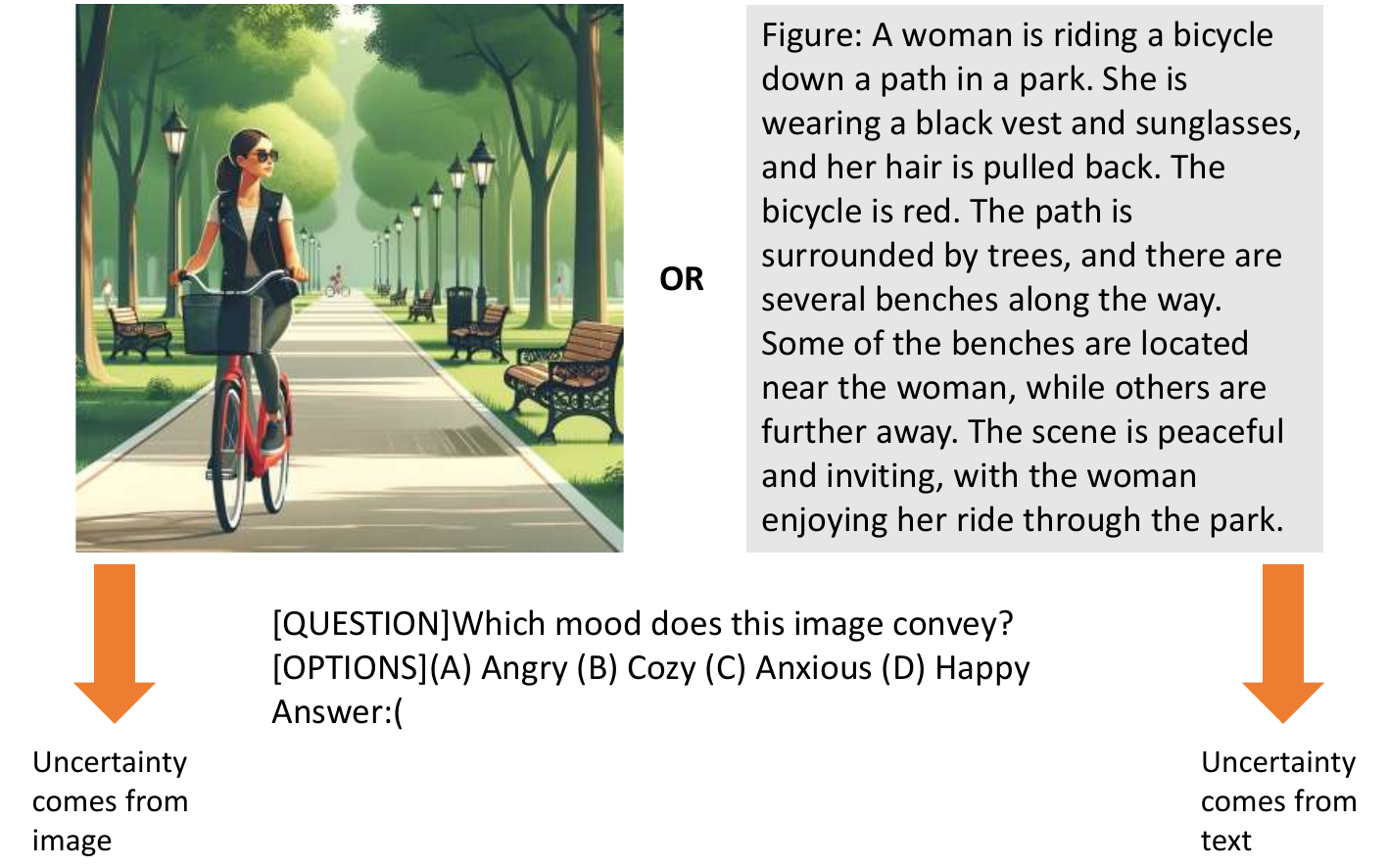}
    \caption{Replaces images with text descriptions, the descriptions are generated by GPT-4V and can accurately describe the images}
    \label{fig:Prompt}
\end{figure}

\subsection{Uncertainty Originating from Different Modalities}
In the VQA task, the model's uncertainty about basic information reflects its uncertainty about the image content.

\textbf{Replacing images with text descriptions.}
Images can be described in text, which introduces another form of uncertainty. For text uncertainty experiments, we generate textual descriptions to replace the images. As shown in Fig.\ref{fig:Prompt}, we use both image and text as the problem's basic information. MLLMs must process these two modalities to answer the question.

Beyond comparing uncertainty across different modalities, we also explore how the model's uncertainty changes across training stages under this setup.

\begin{figure}
    \centering
    \includegraphics[width=1\linewidth]{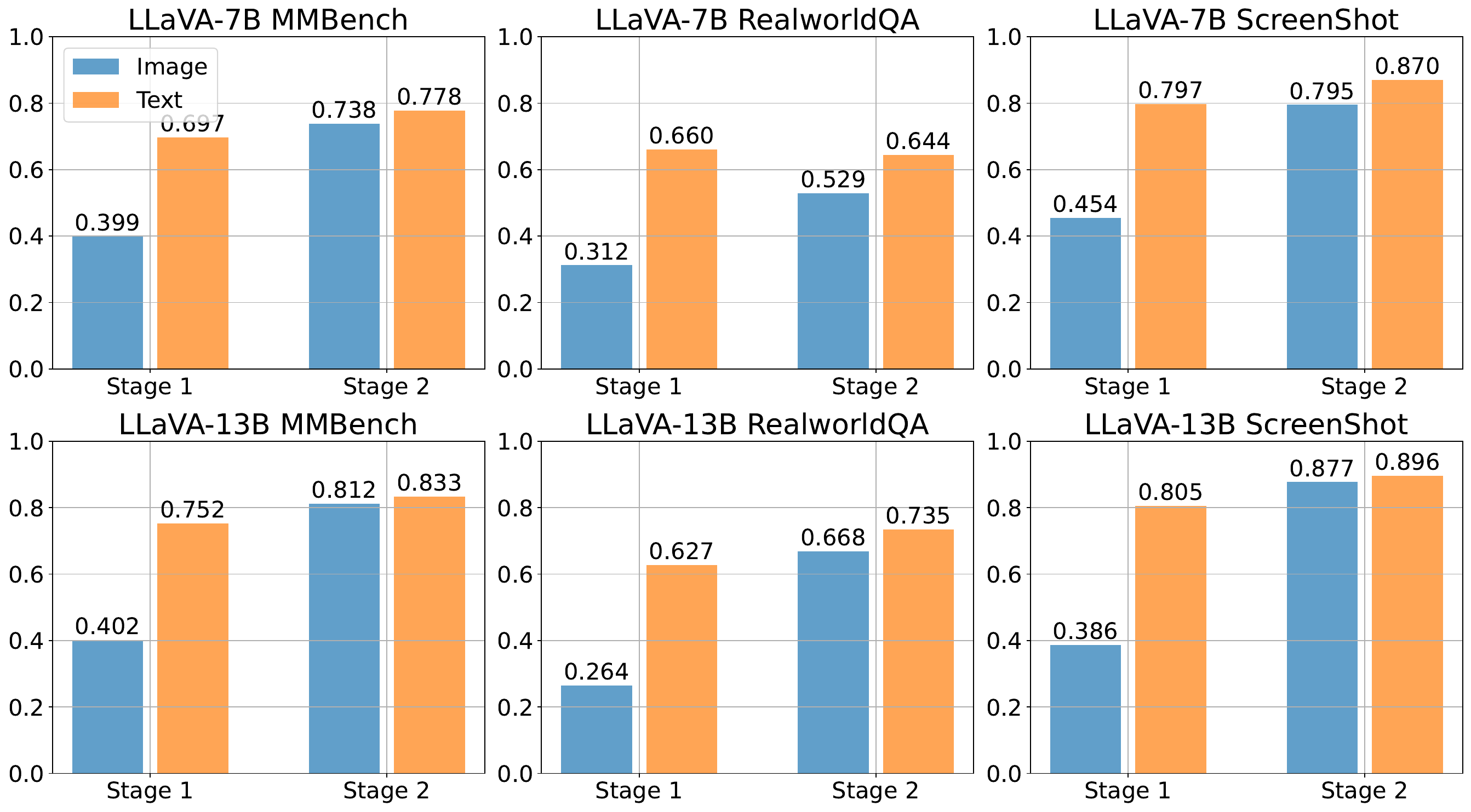}
    \caption{Use logits-based likelihood to quantify model uncertainty, where higher confidence means lower uncertainty. Stage 1 refers to the Pre-Trained MLLMs, while Stage 2 follows visual fine-tuning}
    \label{fig:I&T}
\end{figure}

\textbf{Results and Findings.}
We tested the performance of MLLMs including LLaVA-7B, and LLaVA-13B on their different stages. We use logits-based likelihood to quantify the uncertainty of these models. From Fig.\ref{fig:I&T}, we can see that in the Pre-Trained stage, MLLMs have a much higher confidence in the text than in the image, while after visual fine-tuning, the model has a slightly higher confidence in the text than in the image. This indicates two issues:

\emph{($i$) Compared to text, MLLMs exhibit high uncertainty in information derived from images.}

\emph{($ii$) After visual fine-tuning, MLLMs reduce the uncertainty of the image, which means they will trust the information of the image more.}

\subsection{Integrating Multimodal Information to Reduce Uncertainty}

Humans can boost their confidence in answers by gathering additional information when solving problems. Similarly, we aim to determine if MLLMs can integrate information from different modalities to reduce uncertainty, which is crucial for evaluating a model's performance and reliability. In this section, we extend our uncertainty quantification beyond logits-based likelihood to also include \emph{semantic entropy} \cite{kuhn2023semantic} for open-ended responses. Semantic entropy quantifies uncertainty by clustering a model’s responses based on semantic similarity. If the model’s answers vary significantly in meaning, it indicates high uncertainty. This provides an intuitive measure: the more varied the answers, the less certain the model is. This method bridges language understanding with information theory.

We have created a VQA dataset where images and text both describe a type of organism. The model’s task is to infer the organism from either modality and answer questions about its habits. The dataset is divided into two question types: multiple-choice and open-ended, with model uncertainty quantified using logits-based likelihood and semantic entropy, respectively.

We will add varying levels of Gaussian noise to the images to obscure some of the information they convey. Meanwhile, the textual description will be split into several independent sentences, each describing different characteristics of the organism. We will then progressively add these textual descriptions to the images with different levels of noise. More details in Appendix C. During this process, we will observe how the model's uncertainty changes with the increase in textual information at specific noise levels and compare the uncertainty change curves across different noise levels.

\textbf{Logits-Based vs. Text-Based Uncertainty Quantification.} Logits-based and text-based approaches are two methods for quantifying uncertainty in language models. Logits-based methods are more precise and fine-grained, while text-based methods, which handle discrete outputs, often require multiple inferences. However, text-based methods offer more practical insights for black-box models.

\begin{figure}
    \centering
    \includegraphics[width=1\linewidth]{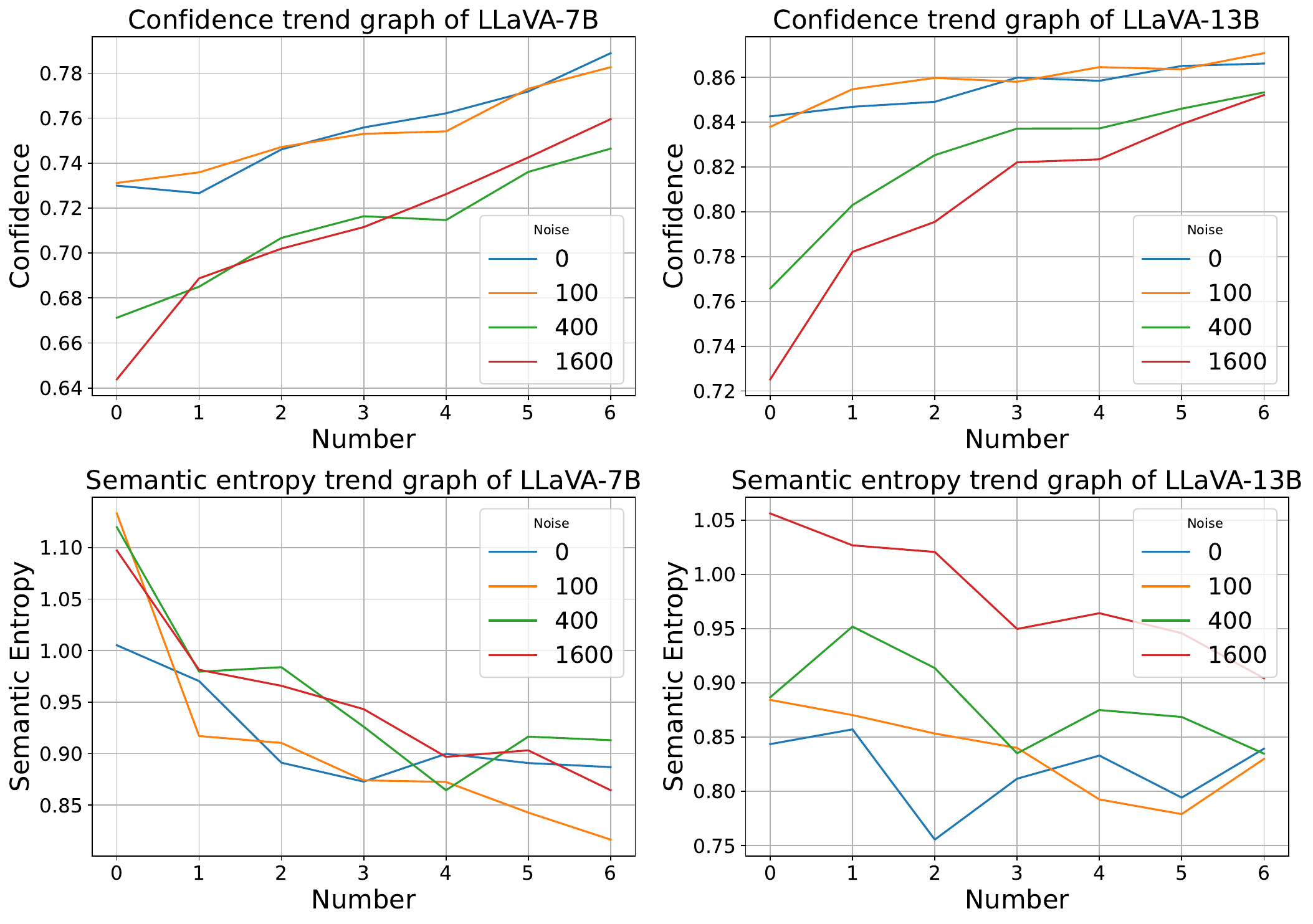}
    \caption{The change in uncertainty of images with different levels of noise as the text description increases. Noise=0 means no noise is added. $NoisyImage$=$Image$+$N(0,Noise)$}
    \label{fig:Trend}
\end{figure}

\textbf{Results and Findings.}
As shown in Fig.\ref{fig:Trend}, We observed that for images with a fixed level of noise, we continuously increase text information, whether it is multiple-choice questions or open answers, resulting in an increase in MLLMs confidence and a decrease in semantic entropy. This indicates that MLLMs can continuously integrate the information of text modalities while obtaining fixed image information, thereby reducing their uncertainty.
In addition, images with higher levels of noise often have higher uncertainty in MLLMs, as can be seen from the points in the image that do not include textual information. With the integration of text information, the model can reduce the uncertainty of images with higher noise more quickly. However, the uncertainty reflected by the model may still be higher than those of clear images. This is because clearer images can also make the model more confident compared to noisy images.
This experiment demonstrates that the model can complement the uncertainty of two modalities, and they will simultaneously affect the uncertainty of MLLMs in the final answer.

\begin{table*}[h!]
\centering
\begin{adjustbox}{width=\textwidth}
\begin{tabular}{lrrrrrlrrrrr}
\toprule
\textbf{Datasets} & \multicolumn{5}{c}{\textbf{LLaVA-7B}} & & \multicolumn{5}{c}{\textbf{LLaVA-13B}} \\
\cmidrule(lr){2-6} \cmidrule(lr){8-12}
 & \textbf{IK-IDK} & \textbf{IDK-IDK} & \textbf{IK-IK} & \textbf{IDK-IK} & \textbf{TRUTHFUL} & & \textbf{IK-IDK} & \textbf{IDK-IDK} & \textbf{IK-IK} & \textbf{IDK-IK} & \textbf{TRUTHFUL} \\
\midrule
MMBench & 0 & 2292 & 2085 & 0 & 47.64\% & & 0 & 1846 & 2529 & 0 & 57.81\% \\
MMBench (Prompting) & 627 & 1665 & 1935 & 150 & 58.53\% & & 151 & 1695 & 2507 & 22 & 60.75\% \\
SEED-Bench & 0 & 9923 & 4310 & 0 & 30.28\% & & 0 & 8671 & 5561 & 0 & 39.07\% \\
SEED-Bench (Prompting) & 2126 & 7797 & 3861 & 449 & 42.06\% & & 273 & 8398 & 5510 & 51 & 40.63\% \\
MobileVQA & 0 & 827 & 44 & 0 & 5.05\% & & 0 & 829 & 42 & 0 & 4.82\% \\
MobileVQA (Prompting) & 526 & 301 & 34 & 10 & 64.29\% & & 622 & 207 & 17 & 25 & 73.36\% \\
PathVQA & 0 & 2610 & 198 & 0 & 7.05\% & & 0 & 2474 & 334 & 0 & 11.89\% \\
PathVQA (Prompting) & 1532 & 1078 & 103 & 95 & 58.23\% & & 1267 & 1207 & 235 & 99 & 53.49\% \\
July24-NewsVQA & 0 & 20968 & / & / & 0\% & & 0 & 20968 & / & / & 0\% \\
July24-NewsVQA (Prompting) & 11990 & 8978 & / & / & 42.82\% & & 5098 & 15870 & / & / & 24.31\% \\
\bottomrule
\end{tabular}
\end{adjustbox}
\caption{In this experiment, MMBench, SEED-Bench, July24-NewsVQA are multiple-choice questions, MobileVQA \cite{MobileVQA}, and PathVQA \cite{he2020pathvqa} are open-ended answer questions. IK-IDK, IDK-IDK, IK-IK, IDK-IK correspond to the number of four domains. \emph{TRUTHFUL} = (\emph{IK-IDK} + \emph{IK-IK}) / (\emph{IK-IDK} + \emph{IDK-IDK} + \emph{IK-IK} + \emph{IDK-IK})}
\label{tab:IDK}
\end{table*}

\begin{table}[h!]
\centering
\resizebox{\columnwidth}{!}{  
\begin{tabular}{lrrr}
\toprule
\textbf{Model} & \textbf{IK-IDK} & \textbf{IDK-IDK}  & \textbf{TRUTHFUL} \\
\midrule
GPT-4o & 61 & 939  & 6.10\% \\
GPT-4o (Prompting) & 697 & 303 & 69.70\% \\
Claude-3-haiku & 133 & 867  & 13.30\% \\
Claude-3-haiku (Prompting) & 677 & 323  & 67.70\% \\
\bottomrule
\end{tabular}
}
\caption{Test two closed source models using 1000 sampled data from the July24-NewsVQA dataset.}
\label{tab:IDK-closed}
\end{table}

\section{CAN MLLMS KNOW WHAT THEY DON’T KNOW?}
LLMs are criticized for their tendency to generate hallucinations. To investigate whether LLMs can recognize when they lack sufficient knowledge on a particular question and express this in natural language, \citet{cheng2024can} constructed a \enquote{I don't know} dataset and tested several models to observe the phenomenon. We referred to their method to construct a VQA dataset for MLLMs

The IDK dataset is divided into two parts: the model-specific dataset segments the existing dataset into what the model \enquote{knows} and \enquote{doesn't know} by having the model repeatedly answer questions. The OOD dataset that is outside the model's training scope, containing questions for which it couldn't possibly know the answers.

\textbf{Construction of dataset:}

\emph{Model-specific dataset:} For a given question, a model answers 10 times. Based on an accuracy threshold (we use 1, meaning all correct), we classify whether the model ``knows" or ``doesn't know." This method segments the existing dataset into IDK and IK categories. For example, in Table~\ref{tab:IDK}, using MMBench to query LLaVA-7B, we segmented the data into IDK and IK, with $2,292$ and $2,085$ entries, respectively.

\emph{OOD dataset:} We constructed the July24-NewsVQA dataset (total of $20968$). This dataset consists of news and pictures from July 2024 (after the models were released) that we scraped, and it was used to create multiple-choice question via GPT-3.5. We assume the model cannot know the OOD answers, so there is no IK portion for this dataset. Some examples of IDK datasets and more construction details are shown in Appendix D. We will opensource this set of the data and the construction details.

We aim to explore whether MLLMs exhibit hallucinations and how well they recognize and express them. To test this, we used the IDK dataset and categorized the results into four domains: \\ \emph{(1)} MLLMs know they don't know (IK-IDK), \\ \emph{(2)} MLLMs don’t know they know (IDK-IDK), \\ \emph{(3)} MLLMs know they know (IK-IK), \\ \emph{(4)} MLLMs don’t know they know (IDK-IK).

\textbf{Results and Findings.}
From Table~\ref{tab:IDK}, we observed that without prompting, MLLMs always provide answers regardless of whether they know the answer or not, indicating an serious overconfidence phenomenon in MLLMs. 
When using prompts such as "If you don’t know the answer, please say..." to encourage MLLMs to handle questions more cautiously, although the models still tend to answer rather than refuse, their accuracy in self-assessment (TRUTHFUL) improves. In addition, compared to multiple-choice questions, open-ended questions have a higher rejection rate, which is intuitive because open-ended answers are more difficult and the accuracy is even lower. This suggests that a well-designed prompt can help alleviate MLLMs' hallucinations to some extent.
For OOD, similar results are observed, even for questions they are unlikely to know, MLLMs still tend to provide answers. However, after being encouraged by prompts, this effect can also generalize to OOD.

To analyze the truthfulness of closed-source multimodal models, we sampled 1,000 instances from the July24-NewsVQA dataset and tested them on two closed-source models: GPT-4o-2024-05-13 and Claude-3-haiku-20240307. Since these models were released before the July24-NewsVQA dataset, they should theoretically be unaware of it.
As shown in Table~\ref{tab:IDK-closed}, when using prompting, larger models has better results(69.7\%, 67.7\%) compared with the smaller models, indicating that the prompting might be more effective to larger models. When prompting is not used, larger models still has better results(6.1\%, 13.3\%) compared with the smaller models(0\%), indicating that the larger models has a better robust to the IDK items. 

\begin{figure}
    \centering
    \includegraphics[width=1\linewidth]{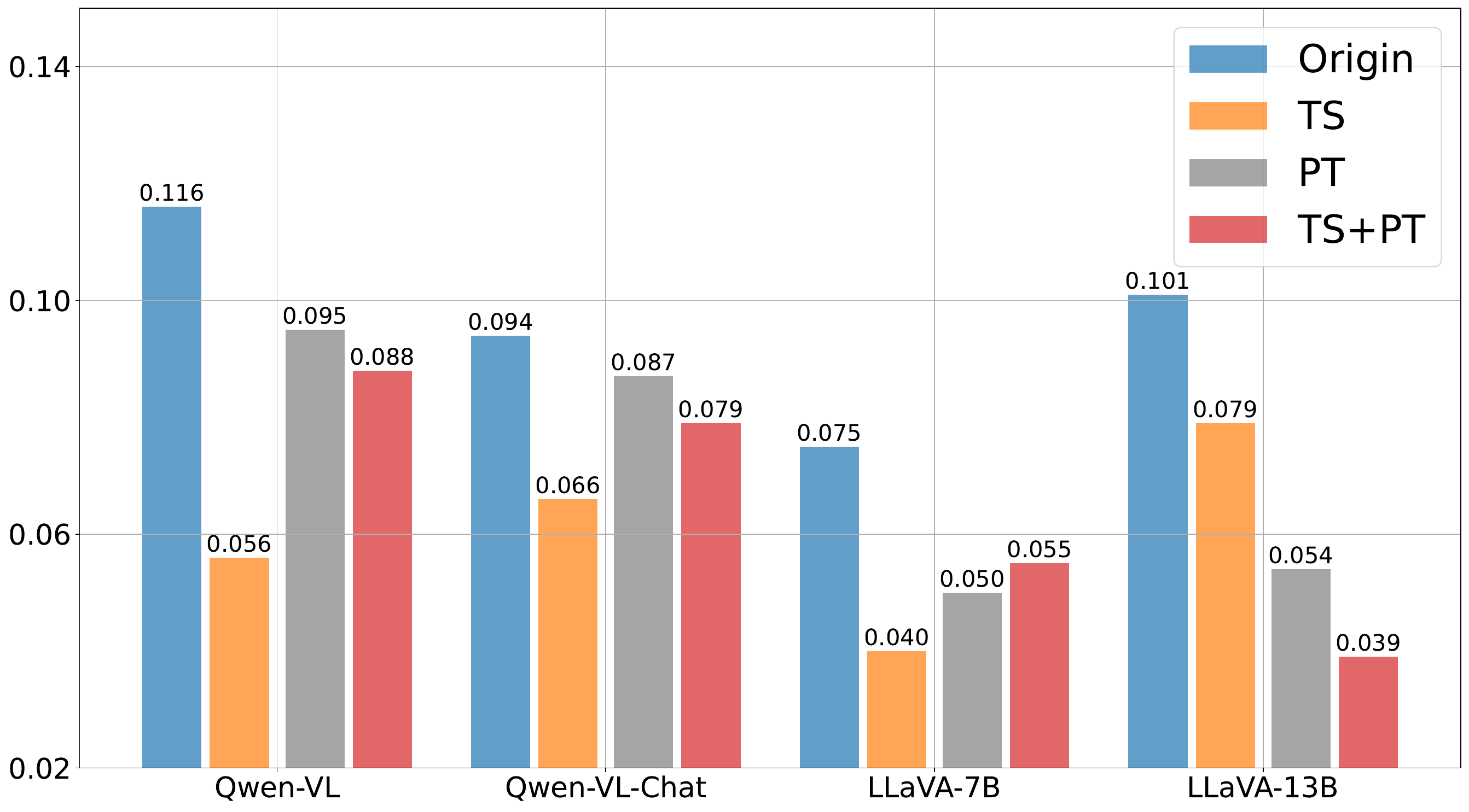}
    \caption{Changes in ECE after calibration for different models tested on MMBench.}
    \label{fig:calibrate}
\end{figure}

\section{CALIBRATION TECHNIQUES FOR MLLMS}
To alleviate miscalibration and enhance the reliability of MLLMs, we propose two techniques specifically for these models.

\subsection{Temperature Scaling with MLLMs’ Predictive Distribution}

Temperature Scaling (TS) is a widely used and effective technique for improving the confidence calibration of neural network classification models~\cite{guo2017calibration}. It works by adjusting the output probabilities of the softmax function, dividing them by a scalar known as the temperature parameter (\(T\)). This parameter plays a crucial role in controlling the smoothness of the resulting probability distribution: higher values of \(T\) produce a smoother, more evenly spread distribution, while lower values result in a sharper, more confident distribution. By adjusting \(T\), TS can help align the model’s confidence with its actual accuracy, making it a valuable method for addressing overconfidence and miscalibration in neural networks.

TS optimizes the temperature \(T\) through the following objective:

\begin{equation}
    \min _{T}-\sum_{i=1}^{M} \sum_{j=1}^{|{y}|} \mathbf{1}_{y_{i}=j} \log \left[\operatorname{softmax}\left(\boldsymbol{l}_{i} / T\right)\right]_{j}.
\end{equation}
Here, \(M\) is the number of samples, \(|y|\) is the number of classes, \(y_i\) is the true label of the \(i\)-th sample, \(\mathbf{1}_{y_i=j}\) is an indicator function, and \(\boldsymbol{l}_i\) are the logits for the \(i\)-th sample. The temperature parameter \(T\) scales the logits \(\boldsymbol{l}_i / T\), controlling the probability distribution's smoothness. When \(T=1\), the output is unchanged; when \(T>1\), the distribution is smoother (less confident); and when \(T<1\), it is sharper (more confident).

Despite its simplicity, TS is highly effective in calibrating MLLMs, especially in models with poor initial calibration, as shown in Fig.\ref{fig:calibrate}.

\begin{algorithm}[H]
\caption{APE for calibration}
\label{alg:ape}
\textbf{Input}: Seed prompts $S$, evaluation function $f$, top prompts $k$, similar prompts $m$, iterations $n$\\
\textbf{Output}: Best prompt
\begin{algorithmic}[1] 
\STATE Initialize $G \leftarrow S$
\FOR{$i = 1$ \textbf{to} $n$}
    \STATE Generate similar prompts for $G$ using GPT: $G_{\text{new}} \leftarrow \bigcup_{p \in G} \text{generate}(p, m)$
    \STATE Evaluate $G_{\text{new}}$: $E \leftarrow \{(p, f(p)) \,|\, p \in G_{\text{new}}\}$
    \STATE Sort $E$ by accuracy bands and ECE
    \STATE Update $G$ with top $k$ prompts from $E$
    \STATE Update best prompt if needed
    \STATE Record current best prompt and top $k$ prompts
\ENDFOR
\STATE \textbf{return} best prompt
\end{algorithmic}
\end{algorithm}

\subsection{Prompt Tuning}

Some researchers have shown that model calibration is influenced by the prompts used \cite{jiang2023calibrating}.  In the previous section, we also observed that prompt can encourage models to better express their uncertainty. Simply adjusting the prompt can help calibrate the model without additional processing. For instance, adding phrases like "This answer might be..." for overly confident models or "This answer must be..." for models lacking confidence can achieve self-calibration.

To achieve self-calibration, we have applied some existing prompt optimization frameworks. The process of prompt tuning has been widely studied, and there are many applicable frameworks that can be adopted, such as APE \cite{ape}, APO \cite{apo} and LongPO \cite{LongPO}. We specifically tuned the prompt suffix, optimizing phrases like "Answer:" to more calibration-friendly versions such as "This answer might be:". The process is outlined in Algorithm~\ref{alg:ape}, which iteratively refines suffixes to improve model calibration (Detail in Appendix E). Prompt tuning can be combined with techniques like TS. Our tests show that in some cases, this combination leads to better calibration results.

\section{CONCLUSIONS}
In this work, we investigated the uncertainty and calibration of MLLMs, focusing on several key areas: the calibration differences between pre-trained and fine-tuned models, the comparison between MLLMs and base LLMs in linguistic tasks, the models' uncertainty when handling text versus images, the integration of uncertainty from both modalities, performance in IDK settings, and calibration techniques.

Our findings reveal that fine-tuned MLLMs do not show significant deterioration in calibration compared to their pre-trained counterparts. Similarly, the multimodal training process, which adjusts the base language model to handle images, does not substantially affect the calibration of linguistic tasks. However, MLLMs still exhibit miscalibration. Specifically, we found that MLLMs display greater uncertainty with image information than with text, and the integration of both modalities affects the overall uncertainty. When tested on the IDK dataset, MLLMs showed a significant overconfidence issue, but strategic prompts improved their ability to self-assess uncertainty. Lastly, we explored calibration techniques like temperature scaling and prompt tuning, demonstrating that these methods effectively enhance MLLM calibration.

The future work of this study includes exploring the integration of the robustness and adversial security of MLLMs~\cite{zhao2024evaluating,wang2023exploring,wang2023iterative}. Additionally, it will explore the uncertainty and calibration of MLLMs in modalities beyond images.

We hope our work provides valuable insights for developing more reliable and robust MLLMs.

\section*{Limitations and Ethical Considerations}
In this study, we introduced several widely used uncertainty quantification methods. However, given the vast number of available techniques and our time constraints, we were unable to explore the impact of other potentially more innovative methods. We also employed standard metrics like ECE for model calibration, but recognize that additional metrics could provide a deeper and more nuanced understanding of the phenomena we observed.

Ethical AI development relies heavily on precise uncertainty quantification and reliability, ensuring that model predictions align with real-world confidence. In high-stakes fields such as healthcare, maintaining human oversight of AI is essential for managing uncertainties, ensuring accountability, and minimizing risks. This oversight is critical not only for mitigating potential errors but also for fostering trust in AI systems. Ethical AI must also prioritize transparency, fairness, and the protection of user rights, ensuring that models behave responsibly under uncertainty and that their limitations are well understood. Balancing AI capabilities with these ethical considerations is key to building systems that are both safe and trusted by society.

\section*{Acknowledgements}
This work is jointly supported by National Natural Science Foundation of China (No. 62306098), the Open Projects Program of State Key Laboratory of Multimodal Artificial Intelligence Systems, the Fundamental Research Funds for the Central Universities (No. JZ2024HGTB0256), the SMP-IDATA Open Youth Fund (No. SMP2023-iData-009) and the Open Project of Anhui Provincial Key Laboratory of Multimodal Cognitive Computation, Anhui University (No. MMC202412).\\
Part of this work was done when the first author was interning at Data Space Research Institute, Hefei, China.

\bibliography{custom}

\clearpage
\appendix
\section*{Appendix A}
Table ~\ref{tab:seed}, ~\ref{tab:bao}, ~\ref{tab:realworldqa}, ~\ref{tab:mathverse}, ~\ref{tab:creature}, ~\ref{tab:screen} show the ECE (MCE and ENCE), accuracy, and confidence of several MLLMs tested on datasets, respectively. These tables are for observing the calibration differences between MLLMs before and after fine-tuning. It can be observed that there is no consistent change in the calibration before and after fine-tuning. But we can still observe the phenomenon of miscalibration.

\begin{table}[h!]
\centering
\scriptsize 
\begin{tabular}{lcccccc}
\toprule
Model & Acc & Conf & ECE & MCE & ENCE \\
\midrule
LLaVA-7B-Stage1 & 0.363 & 0.36 & \textbf{0.075} & 0.438 & \textbf{0.115} \\
LLaVA-7B-Stage2 & 0.636 & 0.645 & 0.105 & \textbf{0.281} & 0.147 \\
\midrule
LLaVA-13B-Stage1 & 0.432 & 0.351 & \textbf{0.101} & 0.294 & \textbf{0.219} \\
LLaVA-13B-Stage2 & 0.636 & 0.776 & 0.139 & \textbf{0.215} & 0.22 \\
\midrule
Qwen-VL & 0.504 & 0.482 & 0.072 & \textbf{0.101} & \textbf{0.133} \\
Qwen-VL-Chat & 0.613 & 0.632 & \textbf{0.071} & 0.46 & 0.245 \\
\bottomrule
\end{tabular}
\caption{ECE (MCE and ENCE), accuracy, and confidence of several MLLMs tested on SEED-Bench}
\label{tab:seed}
\end{table}

\begin{table}[h!]
\centering
\scriptsize 
\begin{tabular}{lcccccc}
\toprule
Model & Acc & Conf & ECE & MCE & ENCE \\
\midrule
LLaVA-7B-Stage1 & 0.412 & 0.357 & \textbf{0.058} & 0.453 & 0.162 \\
LLaVA-7B-Stage2 & 0.606 & 0.686 & 0.081 & \textbf{0.129} & \textbf{0.104} \\
\midrule
LLaVA-13B-Stage1 & 0.577 & 0.369 & 0.207 & 0.445 & 0.314 \\
LLaVA-13B-Stage2 & 0.741 & 0.826 & \textbf{0.096} & \textbf{0.364} & \textbf{0.151} \\
\midrule
Qwen-VL & 0.564 & 0.544 & \textbf{0.113} & 0.192 & 0.136 \\
Qwen-VL-Chat & 0.683 & 0.601 & 0.145 & \textbf{0.174} & \textbf{0.082} \\
\bottomrule
\end{tabular}
\caption{ECE (MCE and ENCE), accuracy, and confidence of several MLLMs tested on BilbaoQA2}
\label{tab:bao}
\end{table}

\begin{table}[h!]
\centering
\scriptsize 
\begin{tabular}{lcccccc}
\toprule
Model & Acc & Conf & ECE & MCE & ENCE \\
\midrule
LLaVA-7B-Stage1 & 0.39 & 0.309 & 0.086 & 0.817 & 0.169 \\
LLaVA-7B-Stage2 & 0.48 & 0.532 & \textbf{0.058} & \textbf{0.288} & \textbf{0.128} \\
\midrule
LLaVA-13B-Stage1 & 0.35 & 0.264 & \textbf{0.091} & 0.244 & 0.253 \\
LLaVA-13B-Stage2 & 0.54 & 0.671 & 0.131 & \textbf{0.236} & \textbf{0.231} \\
\midrule
Qwen-VL & 0.45 & 0.413 & \textbf{0.063} & 0.394 & 0.135 \\
Qwen-VL-Chat & 0.54 & 0.579 & 0.068 & \textbf{0.148} & \textbf{0.113} \\
\bottomrule
\end{tabular}
\caption{ECE (MCE and ENCE), accuracy, and confidence of several MLLMs tested on RealworldQA}
\label{tab:realworldqa}
\end{table}

\begin{table}[h!]
\centering
\scriptsize 
\begin{tabular}{lcccccc}
\toprule
Model & Acc & Conf & ECE & MCE & ENCE \\
\midrule
LLaVA-7B-Stage1 & 0.238 & 0.235 & \textbf{0.059} & \textbf{0.061} & \textbf{0.122} \\
LLaVA-7B-Stage2 & 0.215 & 0.414 & 0.202 & 0.721 & 0.414 \\
\midrule
LLaVA-13B-Stage1 & 0.288 & 0.182 & \textbf{0.106} & \textbf{0.352} & \textbf{0.252} \\
LLaVA-13B-Stage2 & 0.258 & 0.474 & 0.197 & 0.921 & 0.38 \\
\midrule
Qwen-VL & 0.232 & 0.285 & \textbf{0.055} & \textbf{0.443} & \textbf{0.115} \\
Qwen-VL-Chat & 0.314 & 0.453 & 0.146 & 0.959 & 0.253 \\
\bottomrule
\end{tabular}
\caption{ECE (MCE and ENCE), accuracy, and confidence of several MLLMs tested on MathVerse}
\label{tab:mathverse}
\end{table}

\begin{table}[h!]
\centering
\scriptsize 
\begin{tabular}{lcccccc}
\toprule
Model & Acc & Conf & ECE & MCE & ENCE \\
\midrule
LLaVA-7B-Stage1 & 0.635 & 0.497 & 0.162 & 0.703 & 0.23 \\
LLaVA-7B-Stage2 & 0.77 & 0.747 & \textbf{0.092} & \textbf{0.369} & \textbf{0.144} \\
\midrule
LLaVA-13B-Stage1 & 0.53 & 0.379 & 0.157 & 0.316 & 0.267 \\
LLaVA-13B-Stage2 & 0.86 & 0.853 & \textbf{0.04} & \textbf{0.252} & \textbf{0.057} \\
\midrule
Qwen-VL & 0.77 & 0.62 & 0.173 & \textbf{0.374} & 0.217 \\
Qwen-VL-Chat & 0.74 & 0.893 & \textbf{0.158} & 0.699 & \textbf{0.199} \\
\bottomrule
\end{tabular}
\caption{ECE (MCE and ENCE), accuracy, and confidence of several MLLMs tested on Creature}
\label{tab:creature}
\end{table}

\begin{table}[h!]
\centering
\scriptsize 
\begin{tabular}{lcccccc}
\toprule
Model & Acc & Conf & ECE & MCE & ENCE \\
\midrule
LLaVA-7B-Stage1 & 0.6 & 0.454 & 0.145 & \textbf{0.354} & 0.195 \\
LLaVA-7B-Stage2 & 0.79 & 0.795 & \textbf{0.061} & 0.618 & \textbf{0.089} \\
\midrule
LLaVA-13B-Stage1 & 0.64 & 0.386 & 0.275 & 0.456 & 0.412 \\
LLaVA-13B-Stage2 & 0.8 & 0.877 & \textbf{0.077} & \textbf{0.372} & \textbf{0.112} \\
\midrule
Qwen-VL & 0.611 & 0.75 & 0.179 & 0.446 & 0.245 \\
Qwen-VL-Chat & 0.788 & 0.81 & \textbf{0.062} & \textbf{0.374} & \textbf{0.096} \\
\bottomrule
\end{tabular}
\caption{ECE (MCE and ENCE), accuracy, and confidence of several MLLMs tested on ScreenShot}
\label{tab:screen}
\end{table}

\begin{table}[h!]
\centering
\scriptsize 
\begin{tabular}{lcccccc}
\toprule
Model & Acc & Conf & ECE & MCE & ENCE \\
\midrule
Vicuna-7B & 0.700 & 0.687 & 0.033 & \textbf{0.214} & 0.057 \\
LLaVA-7B & 0.785 & 0.794 & \textbf{0.022} & 0.292 & \textbf{0.040} \\
\midrule
Vicuna-13B & 0.831 & 0.826 & \textbf{0.042} & \textbf{0.207} & \textbf{0.056} \\
LLaVA-13B & 0.846 & 0.891 & 0.050 & 0.217 & 0.069 \\
\midrule
LLaMA2-13B-Chat & 0.784 & 0.855 & 0.075 & \textbf{0.181} & 0.097 \\
LLaVA-LLaMA2 & 0.755 & 0.805 & \textbf{0.051} & 0.732 & \textbf{0.078} \\
\midrule
Qwen-7B & 0.844 & 0.763 & \textbf{0.088} & 0.278 & \textbf{0.111} \\
Qwen-VL-Chat & 0.654 & 0.558 & 0.102 & \textbf{0.219} & 0.161 \\
\bottomrule
\end{tabular}
\caption{ECE (MCE and ENCE), accuracy, and confidence of several models tested on ARC}
\label{tab:arc}
\end{table}

\begin{table}[h!]
\centering
\scriptsize 
\begin{tabular}{lcccccc}
\toprule
Model & Acc & Conf & ECE & MCE & ENCE \\
\midrule
Vicuna-7B & 0.490 & 0.559 & 0.069 & \textbf{0.143} & 0.131 \\
LLaVA-7B & 0.626 & 0.658 & \textbf{0.058} & 0.293 & \textbf{0.091} \\
\midrule
Vicuna-13B & 0.620 & 0.684 & \textbf{0.064} & 0.293 & \textbf{0.097} \\
LLaVA-13B & 0.668 & 0.840 & 0.177 & \textbf{0.282} & 0.227 \\
\midrule
LLaMA2-13B-Chat & 0.592 & 0.700 & 0.109 & 0.226 & 0.161 \\
LLaVA-LLaMA2 & 0.596 & 0.675 & \textbf{0.079} & \textbf{0.108} & \textbf{0.133} \\
\midrule
Qwen-7B & 0.664 & 0.605 & 0.058 & \textbf{0.108} & 0.094 \\
Qwen-VL-Chat & 0.626 & 0.641 & \textbf{0.043} & 0.133 & \textbf{0.065} \\
\bottomrule
\end{tabular}
\caption{ECE (MCE and ENCE), accuracy, and confidence of several models tested on OpenbookQA}
\label{tab:openbookqa}
\end{table}

\begin{table}[h!]
\centering
\scriptsize 
\begin{tabular}{lcccccc}
\toprule
Model & Acc & Conf & ECE & MCE & ENCE \\
\midrule
Vicuna-7B & 0.560 & 0.579 & \textbf{0.041} & \textbf{0.148} & \textbf{0.072} \\
LLaVA-7B & 0.708 & 0.852 & 0.152 & 0.707 & 0.188 \\
\midrule
Vicuna-13B & 0.634 & 0.736 & 0.108 & 0.180 & 0.167 \\
LLaVA-13B & 0.624 & 0.682 & \textbf{0.063} & \textbf{0.148} & \textbf{0.084} \\
\midrule
LLaMA2-13B-Chat & 0.600 & 0.700 & 0.102 & 0.294 & 0.142 \\
LLaVA-LLaMA2 & 0.616 & 0.694 & \textbf{0.079} & \textbf{0.148} & \textbf{0.125} \\
\midrule
Qwen-7B & 0.778 & 0.672 & 0.105 & \textbf{0.188} & \textbf{0.137} \\
Qwen-VL-Chat & 0.628 & 0.545 & \textbf{0.097} & 0.204 & 0.165 \\
\bottomrule
\end{tabular}
\caption{ECE (MCE and ENCE), accuracy, and confidence of several models tested on CommonsenseQA}
\label{tab:commonsense_qa}
\end{table}

\begin{table}[h!]
\centering
\scriptsize 
\begin{tabular}{lcccccc}
\toprule
Model & Acc & Conf & ECE & MCE & ENCE \\
\midrule
Vicuna-7B & 0.632 & 0.705 & 0.075 & 0.208 & 0.103 \\
LLaVA-7B & 0.740 & 0.755 & \textbf{0.056} & \textbf{0.118} & \textbf{0.078} \\
\midrule
Vicuna-13B & 0.704 & 0.755 & \textbf{0.066} & \textbf{0.286} & \textbf{0.096} \\
LLaVA-13B & 0.708 & 0.836 & 0.136 & 0.288 & 0.174 \\
\midrule
LLaMA2-13B-Chat & 0.696 & 0.786 & 0.090 & 0.191 & 0.137 \\
LLaVA-LLaMA2 & 0.682 & 0.734 & \textbf{0.066} & \textbf{0.110} & \textbf{0.101} \\
\midrule
Qwen-7B & 0.806 & 0.689 & 0.116 & 0.204 & 0.151 \\
Qwen-VL-Chat & 0.637 & 0.623 & \textbf{0.047} & \textbf{0.108} & \textbf{0.086} \\
\bottomrule
\end{tabular}
\caption{ECE (MCE and ENCE), accuracy, and confidence of several models tested on RACE}
\label{tab:race}
\end{table}

\section*{Appendix B}
Table ~\ref{tab:arc}, ~\ref{tab:openbookqa}, ~\ref{tab:commonsense_qa}, ~\ref{tab:race} show the accuracy and ECE (MCE and ENCE) of several models tested on CommonsenceQA and RACE, respectively. After training the visual modality, the calibration of the model for linguistic tasks remained relatively stable, But we can still observe the phenomenon of miscalibration.

\section*{Appendix C}
\begin{figure}
    \centering
    \includegraphics[width=1\linewidth]{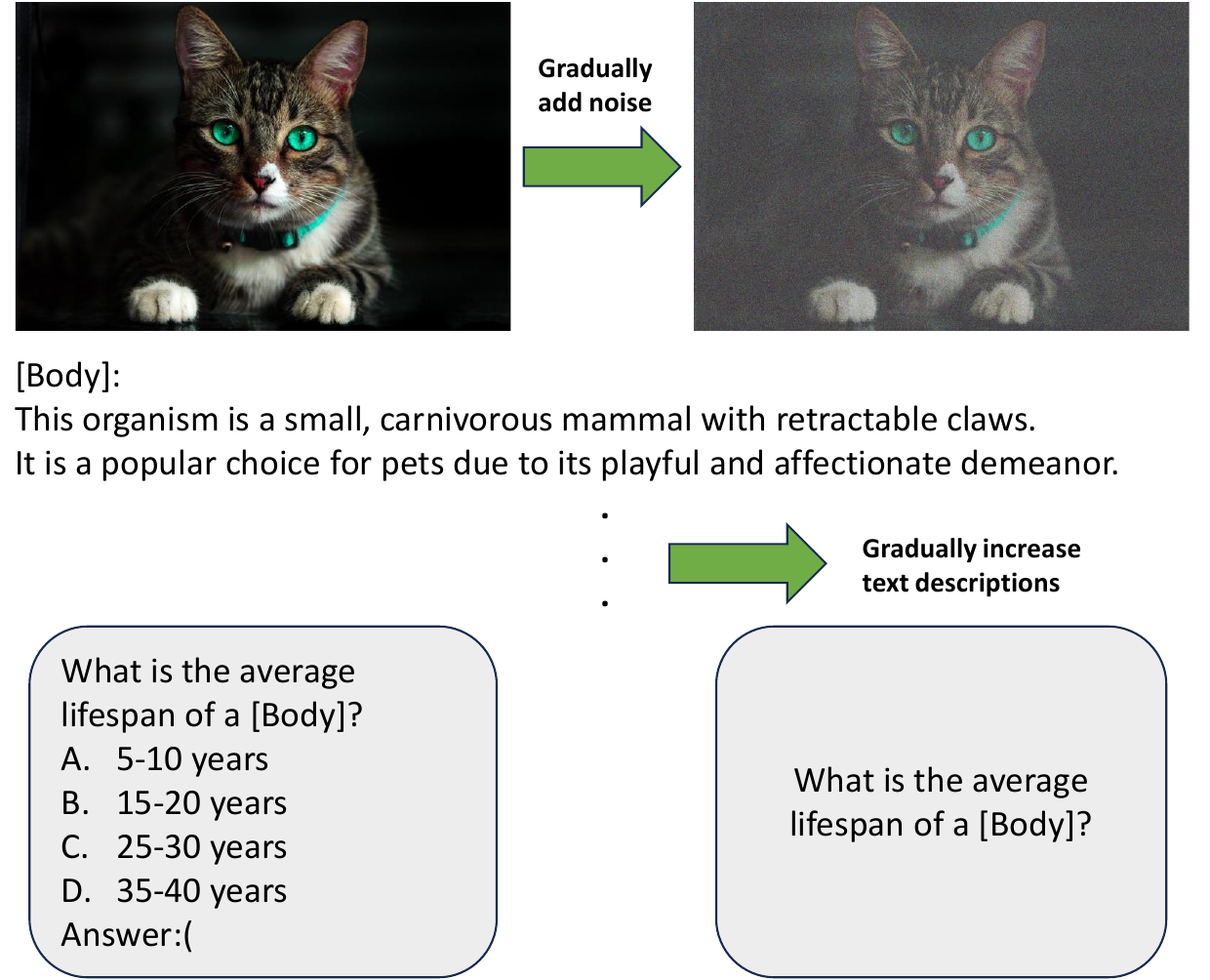}
    \caption{Gradually add text descriptions on images with different levels of noise, and observe the changes in uncertainty of information integration models for the two modalities}
    \label{fig:Cat}
\end{figure}

We further explain how to continuously integrate information from different modalities to observe changes in uncertainty. Fig.\ref{fig:Cat} shows the overall method

Specifically:

\begin{enumerate}
    \item Add Gaussian noise with different standard deviations \( \sigma_i \) to the original image \( I \), obtaining the set of noisy images \( \{I_{\sigma_1}, I_{\sigma_2}, \ldots, I_{\sigma_n}\} \).
    \item Split the textual description into several independent sentences \( S = \{S_1, S_2, \ldots, S_m\} \).
    \item For each noise level \( \sigma_i \), progressively add the textual descriptions \( S_{\leq k} \), where \( S_{\leq k} = S_1 + S_2 + \ldots + S_k \), and calculate the model's uncertainty \( U(I_{\sigma_i}, S_{\leq k}) \).
    \item Plot and analyze the change in model uncertainty \( U \) with increasing textual descriptions \( S_{\leq k} \) for different noise levels \( \sigma_i \).
\end{enumerate}

\section*{Appendix D}

Here are some examples of IDK datasets we constructed.

\begin{figure}
    \centering
    \includegraphics[width=1\linewidth]{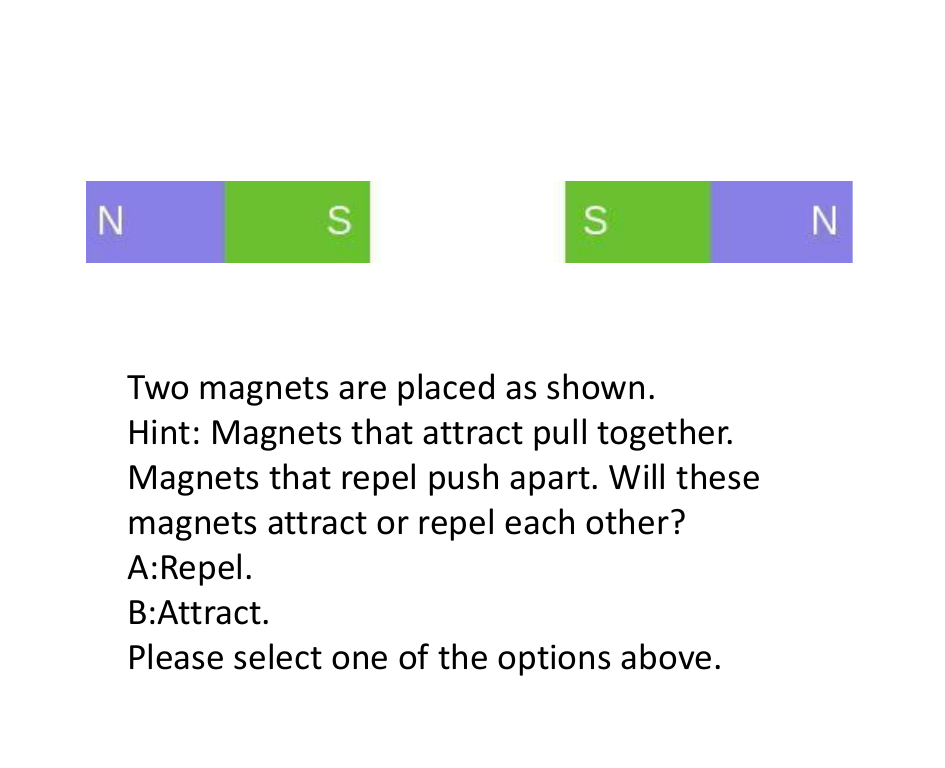}
    \caption{This question was answered 10 times by LLaVA-7B but was not answered correctly each time. We believe this model does not know the answer.}
    \label{fig:example3}
\end{figure}

\begin{figure*}
    \centering
    \includegraphics[width=\textwidth]{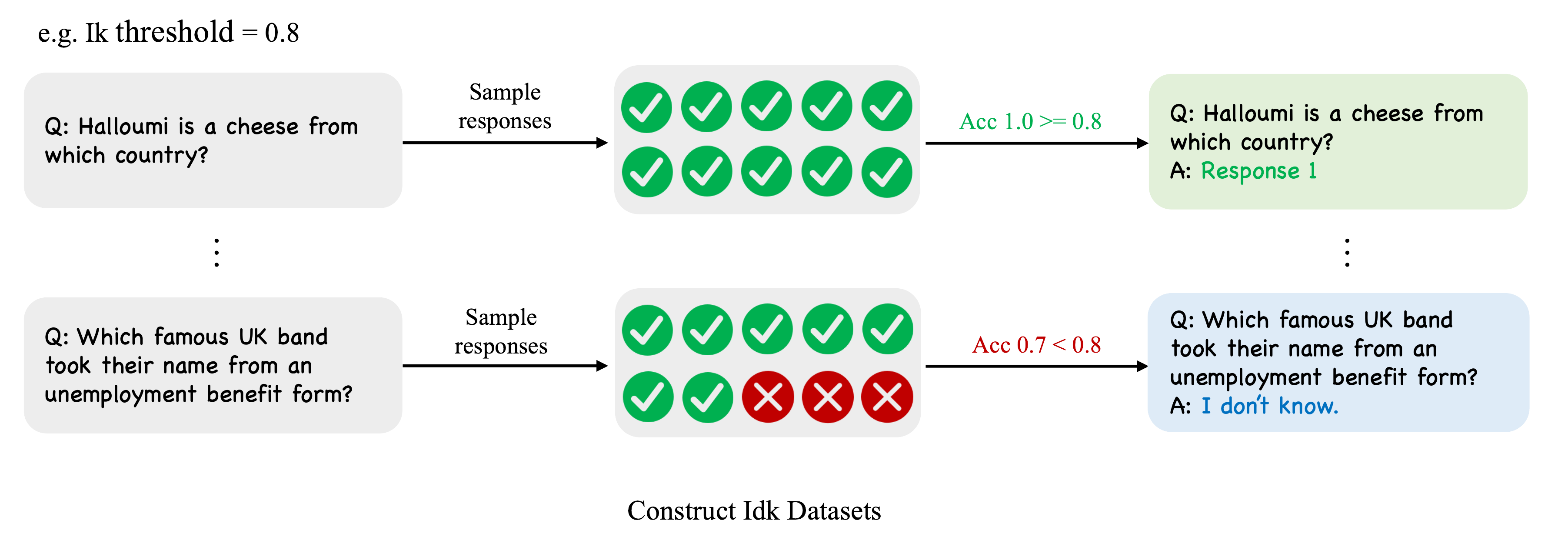}
    \caption{This figure comes from \cite{cheng2024can}. We have the model answer multiple times and determine whether it knows the answer to the question based on a pre-set threshold.}
    \label{fig:idk}
\end{figure*}

\textbf{model-specific dataset:} Fig.\ref{fig:example3} shows LLavVA-7B repeatedly answering the same question 10 times in the MMBench dataset, but failing to get all answers correct. LLaVA-7B uses a sampling strategy with a temperature of 1 and top-p set to 0.95. Fig.\ref{fig:idk} shows the process of constructing the IDK dataset. When the accuracy of multiple answers is less than the threshold, it is considered that the model does not know. Specifically:

\emph{($i$) For a single question-answer data point, have the model answer 10 times and record the accuracy.}

\emph{($ii$) For data items with accuracy below a specified threshold, label them as 'I do not Know'.}

This dataset is selected from previously public datasets by choosing items that the model does not know, and is used to evaluate the performance of MLLMs on questions they do not know. Our contribution lies in applying \citet{cheng2024can}'s construction method for LLMs to MLLMs and constructing a VQA dataset.

\begin{figure}
    \centering
    \includegraphics[width=1\linewidth]{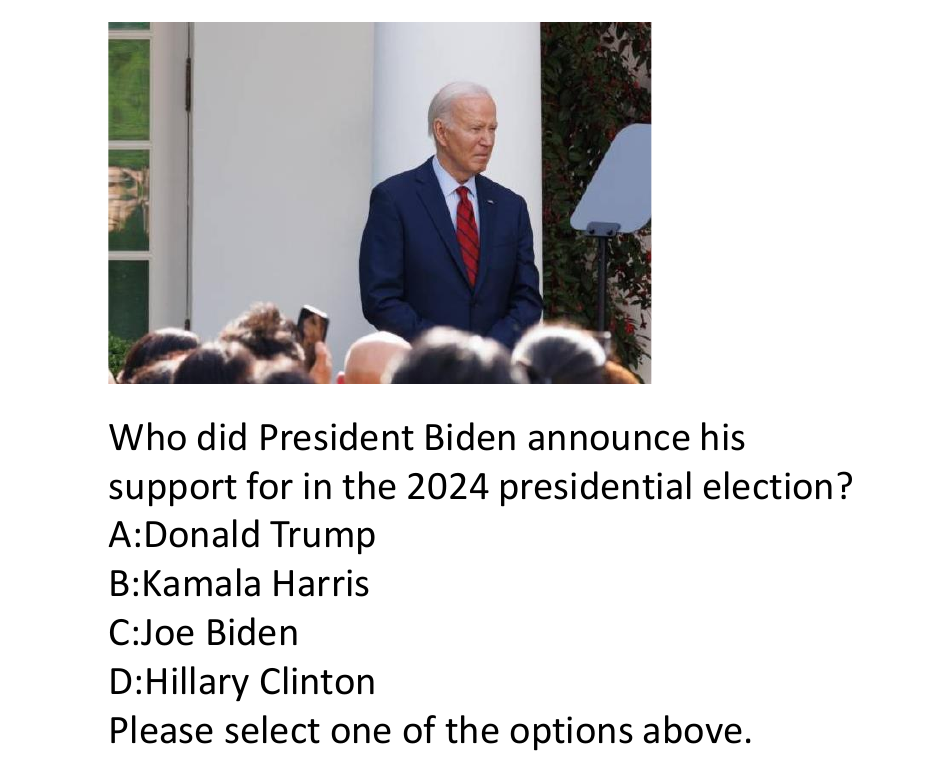}
    \caption{The models we tested were trained before July 2024, so they absolutely cannot know who Biden announced his support for in the 2024 election.}
    \label{fig:example1}
\end{figure}

\begin{figure}
    \centering
    \includegraphics[width=1\linewidth]{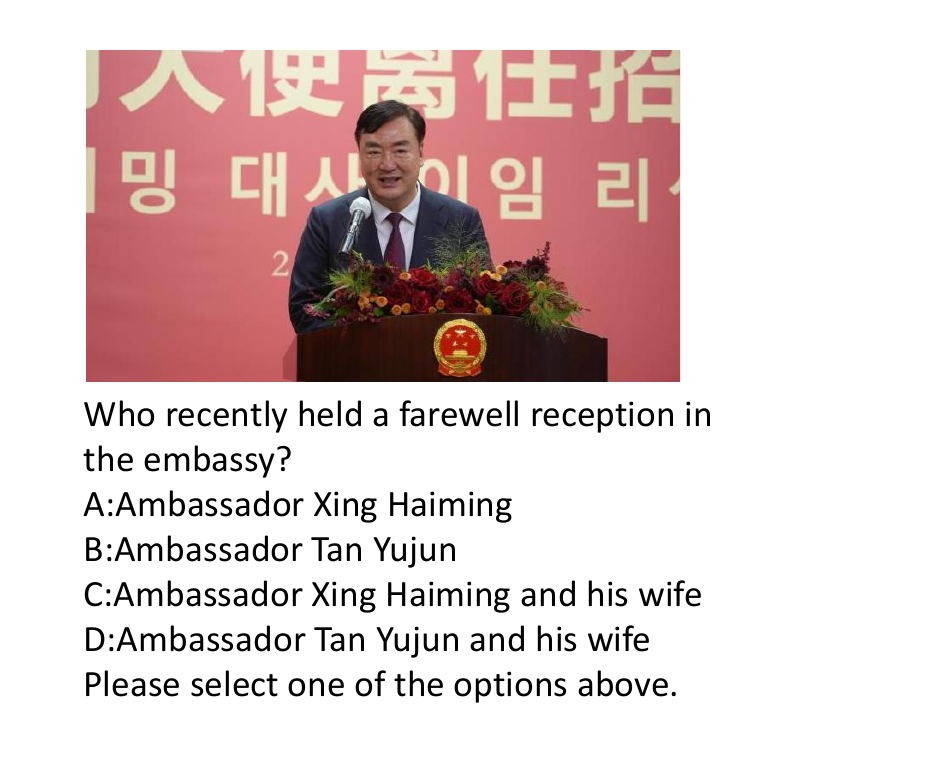}
    \caption{The models we tested were trained before July 2024, so they absolutely cannot know who recently held a farewell reception in the embassy.}
    \label{fig:example2}
\end{figure}

\begin{figure*}
    \centering
    \includegraphics[width=\textwidth]{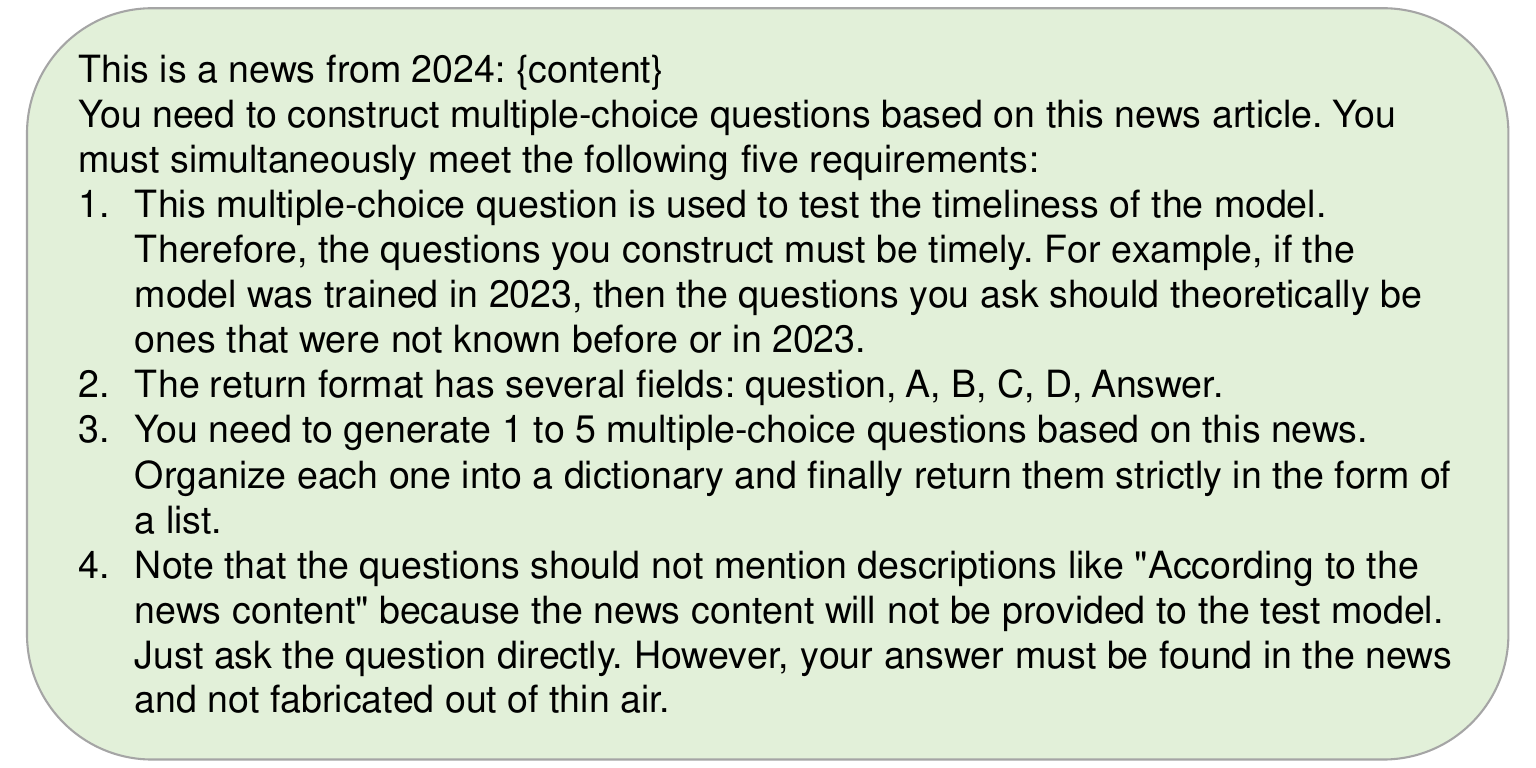}
    \caption{This figure shows the prompt for constructing multiple-choice questions from crawled news by GPT.}
    \label{fig:OOD_prompt}
\end{figure*}

\textbf{OOD dataset:} Fig.\ref{fig:example1} and Fig.\ref{fig:example2} show the news from July 24 that we crawled and constructed the question using GPT-3.5 (prompt is showed in Fig.\ref{fig:OOD_prompt}). We scraped the news from \url{https://www.chinanews.com.cn}. Specifically:

\emph{($i$) Scrape the article contents and accompanying images from news websites for July 2024.}

\emph{($ii$) Provide the article content to GPT-3.5 and use the aforementioned prompt to generate several multiple-choice questions.}

Our contribution lies in proposing a method to construct an OOD (Out-of-Distribution) dataset for MLLMs by transforming the latest news article contents and accompanying images through language models. We provide a dataset used to evaluate the performance of MLLMs on unknown questions. It contains 6,774 images and 20,968 multiple-choice questions (each image corresponds to multiple question-and-answer items). The dataset contains 7 fields, namely question, option A, option B, option C, option D, answer, and image path.

\section*{Appendix E}
The process, outlined in Algorithm 1, begins with a set of initial suffixes. In each iteration, GPT-3.5 generates similar suffixes, which are evaluated for accuracy and ECE. Suffixes are then grouped by accuracy and sorted by ECE in ascending order to prioritize well-calibrated options. The top \(k\) suffixes are selected for the next iteration. After the specified iterations, the algorithm returns the best-performing suffix.

\end{document}